\documentclass[10pt,conference]{IEEEtran} 
\usepackage[english]{babel}

\usepackage{times}
\usepackage{caption}
\usepackage{graphicx}
\newcommand{\fig}[4][htbp]{
\begin{figure}[#1]
\centering%
\includegraphics[#2]{#3}
\caption[#4]{#4.}\label{fig:#3}
\end{figure}}
\usepackage{etoolbox}
\usepackage{siunitx}
\usepackage{booktabs}
\usepackage{color}

\usepackage[pdftex, pagebackref=false, colorlinks=true, linkcolor=black, 
citecolor=black, filecolor=black, urlcolor=black, plainpages=false]{hyperref}
\usepackage{url}
\hypersetup{%
pdfauthor={Nuno A. C. Henriques},
pdftitle={SensAI+Expanse Adaptation on Human Behaviour Towards Emotional 
Valence Prediction},
pdfkeywords={emotional valence prediction, context adaptation, memory, HAI},
pdfproducer={nunoachenriques.net}}

\usepackage{glossaries}
\setacronymstyle{long-short}
\newacronym{anr}{ANR}{Application Not Responding}
\newacronym{automl}{AutoML}{Automated Machine Learning}
\newacronym{gps}{GPS}{Global Positioning System}
\newacronym{hai}{HAI}{Human-Agent Interaction}
\newacronym{mcc}{MCC}{Matthews Correlation Coefficient}
\newacronym{weird}{WEIRD}{Western, Educated, Industrialized, Rich, and 
Democratic}
\newacronym{xai}{XAI}{Explainable Artificial Intelligence}

\def\dummyname{Dummy}
\def\lrname{Logistic Regression}
\def\xgboostname{Extreme Gradient Boosting}
\def\nnname{TensorFlow Keras MLP}
\def\nunoachenriques{\href{https://nunoachenriques.net/}{Nuno A. C. Henriques}}

\def\android{\href{https://developer.android.com/}{Android}}
\def\pvalue{$p\ \textnormal{value}$}
\def\sensaigit{https://gitlab.com/nunoachenriques/sensei}
\def\sensai{\href{\sensaigit}{SensAI}}
\def\sensaiexpansegit{https://gitlab.com/nunoachenriques/sensei-expanse}
\def\sensaiexpanse{\href{\sensaiexpansegit}{SensAI Expanse}}
\def\expanse{\href{\sensaiexpansegit}{Expanse}}
\def\sensaiplusexpanse{\sensai+\expanse}
\def\sklearnurl{https://scikit-learn.org/}
\def\sklearnpkg{\texttt{\href{\sklearnurl}{scikit-learn}}}
\def\twitter{\href{https://twitter.com/}{Twitter}}

\begin{document}
\pagenumbering{gobble}
\title{\textbf{\Large \sensaiplusexpanse\\[-1.5ex] Adaptation on Human 
Behaviour Towards Emotional Valence Prediction}\\[0.2ex]}
\author{\IEEEauthorblockN{~\\[-0.4ex]\large 
\nunoachenriques\\[0.3ex]\normalsize}
\IEEEauthorblockA{BioISI\\
Faculdade de Ci\^{e}ncias\\
Universidade de Lisboa\\
Portugal\\
{\tt nach@edu.ulisboa.pt}}
\and
\IEEEauthorblockN{~\\[-0.4ex]\large Helder Coelho\\[0.3ex]\normalsize}
\IEEEauthorblockA{BioISI\\
Faculdade de Ci\^{e}ncias\\
Universidade de Lisboa\\
Portugal\\
{\tt hcoelho@di.fc.ul.pt}}
\and
\IEEEauthorblockN{~\\[-0.4ex]\large Leonel Garcia-Marques\\[0.3ex]\normalsize}
\IEEEauthorblockA{CICPSI\\
Faculdade de Psicologia\\
Universidade de Lisboa\\
Portugal\\
{\tt garcia\_marques@sapo.pt}}}
\maketitle
\begin{abstract}
An agent, artificial or human, must be continuously adjusting its behaviour in
order to thrive in a more or less demanding environment. An artificial agent
with the ability to predict human emotional valence in a geospatial and temporal
context requires proper adaptation to its mobile device environment with
resource consumption strict restrictions (e.g., power from battery). The
developed distributed system includes a mobile device embodied agent (\sensai)
plus Cloud-expanded (\expanse) cognition and memory resources. The system is
designed with several adaptive mechanisms in the best effort for the agent to
cope with its interacting humans and to be resilient on collecting data for
machine learning towards prediction. These mechanisms encompass homeostatic-like
adjustments, such as auto recovering from an unexpected failure in the mobile
device, forgetting repeated data to save local memory, adjusting actions to a
proper moment (e.g., notify only when human is interacting), and the \expanse\
complementary learning algorithms' parameters with auto adjustments. Regarding
emotional valence prediction performance, results from a comparison study
between state-of-the-art algorithms revealed \xgboostname\ on average the best
model for prediction with efficient energy use, and explainable using feature
importance inspection. Therefore, this work contributes with a smartphone
sensing-based system, distributed in the Cloud, robust to unexpected behaviours
from humans and the environment, able to predict emotional valence states with
very good performance.
\end{abstract}


\section{Introduction}

The scientific evidence of epigenetics reveal on/off mechanisms inside
chromosomes of human agents and reinforces the importance of any entity
continuous adaptation to its environment. Additionally, some natural entities
such as human individuals with self-consciousness and emotion-driven cognition
developed a bond between the evolutionary way of emotions and their supporting
physical structure as proposed by Damásio \cite{Damasio2010EN}. In a sense, it
is clear that an agent's behaviour will not develop independently of the
environment and that its affective states are paramount in the adjustment.
Further, a developed behaviour may be the result of an ongoing, bidirectional
interchange between inherited traits (e.g., parameter initial value) and the
environment (e.g., data collected from an interacting human). Therefore, it may
be envisioned an artificial agent adjusting empathetically towards the
interacting human current behaviour and affective state
\cite{Picard1997}\cite{Castellano2010}. The concept of
empathy~\cite{Stueber2014} may be used as a starting point for social glue
bringing better interaction, communication and mutual helping. \Gls{hai} should
be based on the way each entity perceives contact, together with the perception
of human's affective states in a multimodal approach
\cite{Tavabi2019}\cite{Tsiourti2018}. Hence, the affect sensing using  wearable
or mobile devices, such as a smartphone seems appropriate. The American College
of Medical Informatics (ACMI) has already envisaged this path. In the 1998
Scientific Symposium, one of the informatics challenges for the next 10 years
was ``Monitor the developments in emerging wearable computers and sensors ---
possibly even implantable ones --- for their potential contribution to a
personal health record and status monitoring''~\cite{Greenes1998}. Twenty years
have passed since this awakening for the mobile device as a sensing tool.
Smartphone sensing for behavioural research is thriving with active discussions
\cite{Denecke2019}\cite{Felix2019} including exploration on correlates between
sensors' data and depressive symptom severity \cite{Saeb2015}.

This paper describes the \sensaiplusexpanse\ system and its adaptive mechanisms
towards emotional valence prediction ability on humans. The individuals may be
diverse in behaviour, age, gender and place of origin. Accordingly, the
developed system encompass a mobile device embodied agent \sensai\ and its
Cloud-expanded (\expanse) cognition and memory resources. \sensai\ collects data
from several sources including (a) device sensors, such as \gls{gps} and
accelerometer; (b) current timestamp in user calendar; and (c) available text
writings from in-application diary and social network posts (\twitter\ status).
These written texts in (c) will be subjected to sentiment analysis
\cite{Hutto2014} in order to collect emotional valence from this modality
source. The ground truth is obtained from the user when reporting about current
sentiment (positive, neutral, negative). On the other hand, the artificial agent
will be subjected to a simple adaptive process by means of interaction with
humans. An empathy score value is presented during this interaction. The score
decays over time, it also changes with some factors, such as the frequency of
human reporting. This visual adaptive metric should be perceived by the human as
current human-agent empathetic score. The \expanse\ complementary resources
comprise several heuristics and algorithms, such as unsupervised location
clustering parameters auto discovery and supervised learning hyperparameters
auto tuning. These are continuously adapting to the data set of each human
entity. Further, preliminary results from a running study with the agent in the
wild, publicly available for installation, are presented. This methodology
contributes to avoid a well-known \gls{weird} society bias in research studies
involving human subjects exclusively from academia. Moreover, performance
results of a comparison study between state-of-the-art machine learning
algorithms are presented and a model is elected as the best for future studies.

The first section introduced the purpose of this investigation and the work done
so far. Next, Section \ref{sec:adaptive} will describe the mechanisms in place
for the developed mobile agent system adaptive capabilities. Section
\ref{sec:study} describes the research study including the followed method and
the achieved results. Finally, Section \ref{sec:conclusion} summarises the
outcomes and presents a future perspective.

\section{\sensaiplusexpanse\ Adaptive Mechanisms}\label{sec:adaptive}

This section describes the adaptive mechanisms in place for the developed
\sensaiplusexpanse\ as a distributed, fault-tolerant, mobile, and Cloud-based
system from scratch. The platform is used as a research tool for continuously,
online, gather and process sensory data. 
Figure~\ref{fig:sensai+expanse---concept-flow} depicts the general data flow 
between the \sensai\ agent and its expanded resources. The collected data from 
mobile sensors and \gls{hai} is processed and stored locally. Additionally, 
data is periodically synced in order to feed the learning process and 
prediction service.

\fig{scale=0.85}{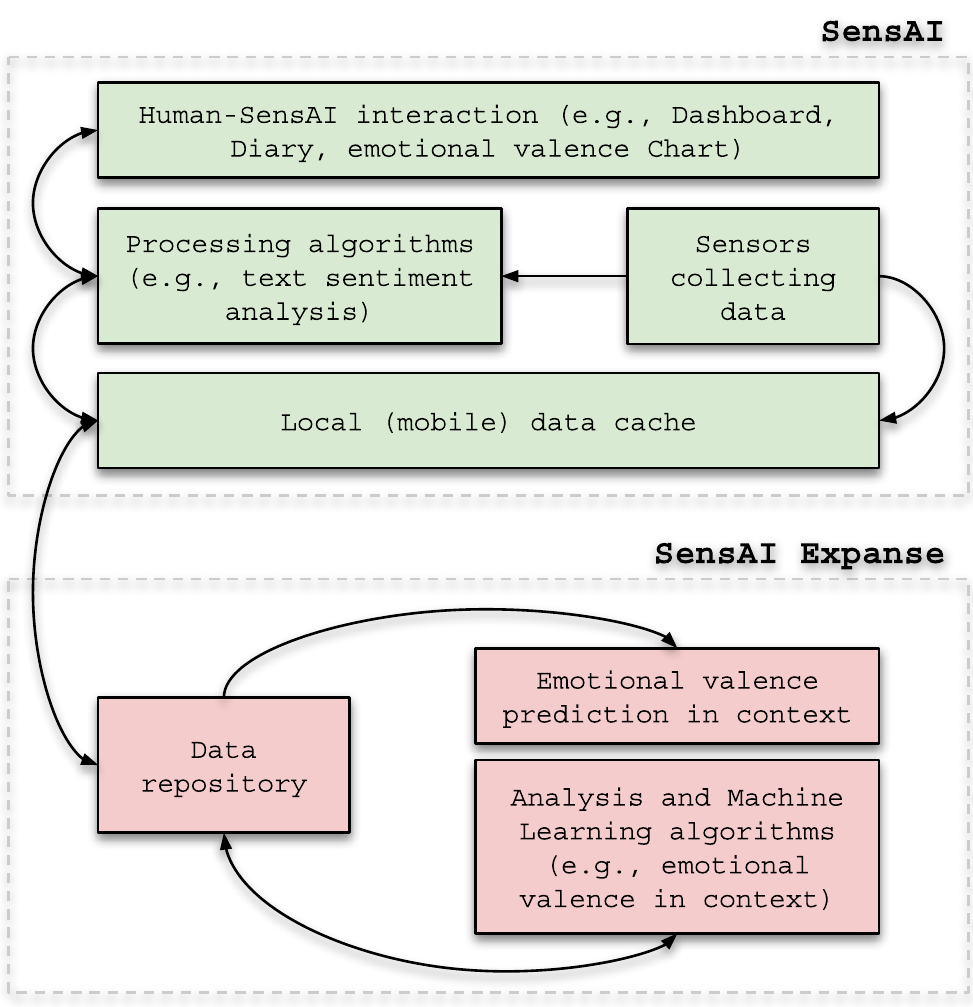}{\sensaiplusexpanse\ general 
data flow}

The general \gls{hai} is initially restricted by its parameter values which
drives \sensai. This behaviour may be influenced by the agent's context along
the interaction timeline and changes may emerge as adjustment details.
Complementary, \sensaiexpanse\ contains a myriad of adaptive mechanisms
regarding collected data from human behaviour. These actions work towards
\gls{automl} and efficient prediction. The emotional valence ground truth values
used for prediction performance measurement are reported by humans. The main
interface includes three emoticons as depicted in
Figure~\ref{fig:sensai---main-ui}. Moreover, this mechanism is robust to
interaction bias, such as high-frequency repeated button (emoticon) clicks. 
Also,
on cases of mistaken valence promptly corrected by an additional hit on a
different emoticon. It includes a simple yet effective heuristic of accounting
only for the last hit during a defined time interval. All these actions are
contextualised, i.e., the location and timestamp of the event are collected.

\fig{scale=0.9}{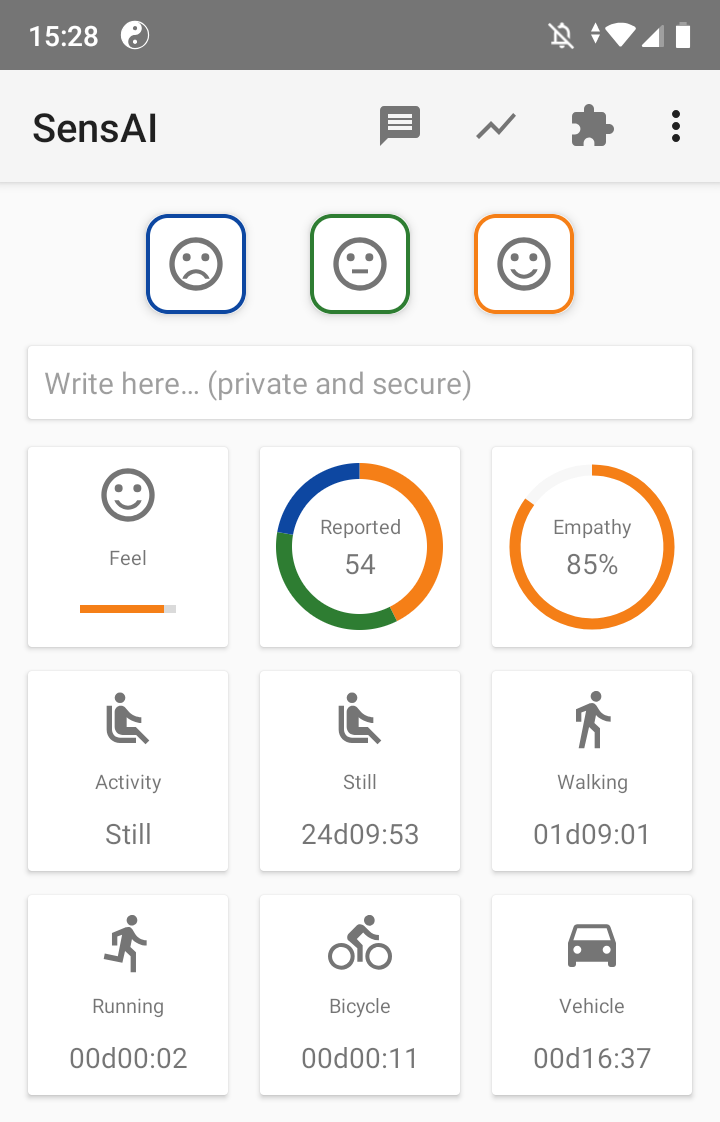}{\sensai\ main user interface and system 
notification bar}

\sensai\ has two ways of collecting data by doing it (a) passively using several
sensors, such as accelerometer and \gls{gps}; and (b) actively by interacting
with the human using display notifications and buttons for emotional valence
reporting. Moreover, \sensai\ displays information about the human physical
activity aggregated time by each recognisable type (e.g., running), current
emotional valence status, self-report statistics, and agent empathy score as
depicted in Figure~\ref{fig:sensai---main-ui}. Additional displays are available
with (a) sentiment chart with the chronology of reporting and messaging
emotional valence values; (b) private diary for writing messages to self where
text is subjected to sentiment analysis \cite{Hutto2014}, including a \sensai\
report of current averaged emotional valence status and physical activity every
three or so hours; and (c) several statistics about sensors event count, last
\expanse\ data sync, and data collecting uptime.

\subsection{\sensai}

The mobile device embodied agent has several mechanisms in place for specific
adjustments. These workings are included in different modules. Each one of
those is autonomously managed although orchestrated by \texttt{Homeostasis}
module with periodic health checks. \texttt{SenseiStartStop} is a fail-safe last
resort to deal with device start/stop and also unexpected \sensai\ failures such
as asynchronous illegal states causing the application to crash, i.e., be
removed from running state. Activity, service and special modules are
instantiated objects from code developed classes. Some run on demand, others
periodically, as services and activities on dedicated threads or the main user
interface thread. The relevant activities and services for \sensai\ embodiment
are described below.

\texttt{Homeostasis} is paramount to guarantee some tolerance to failures and
keep the agent in a good health, it is a scheduled, service designed
to regulate the embodiment. Every run checks for critical aspects, such as
database health and data feed. It takes proper actions to solve some common
failures, such as sensor data \texttt{Feed} not running. Moreover, adapt itself
to the interaction state, i.e., if at rest then database optimisations and fix
actions may run, conversely, updating notifications, such as empathy level only
happen when the human user is paying attention. This mechanism prevents
potentially disturbing events, such as too frequent device's screen awaking just
for an empathy value adjustment. The homeostasis-like solution for the \sensai\
application is complemented with \texttt{SenseiStartStop} required to protect
and guarantee \texttt{Homeostasis} service to run as expected.

\texttt{SenseiStartStop} is a system event receiver to assure persistence and 
robustness against the device failures and reboots. It does a system 
registration at \sensai\ first start to be called on device boot and on 
application upgrade dealing with those special states. This registration also 
signals \android\ operating system to revive \sensai\ in case of unexpected 
crash and removal from running state.

\texttt{Feed} is a started service running autonomously in the background.
Several other services run on demand in an adaptation to save mobile device
resources consumption, such as battery. This module encompasses and manages all
data collecting from sensors, such as \android-device hardware types (e.g.,
accelerometer). Moreover, a balanced data acquisition rhythm, such as
$active=2s$, $inactive=8s$, $f=1/5Hz$, and $D=20\%$ is devised and in place for
relevant data to be acquired without draining too much power. This rhythm as
well as other thresholds may be subjected to automatic adaptation in the future.
Furthermore, Figure~\ref{fig:sensai---empathy-notification} depicts a persistent
notification message which includes empathy level adjustment in a progress bar
triggered by emotional valence reports (using emoticon buttons). Also, a
dashboard is available with relevant information including empathy level. An
active \sensai\ main user interface dashboard was already depicted in
Figure~\ref{fig:sensai---main-ui}.

\fig{scale=1.0}{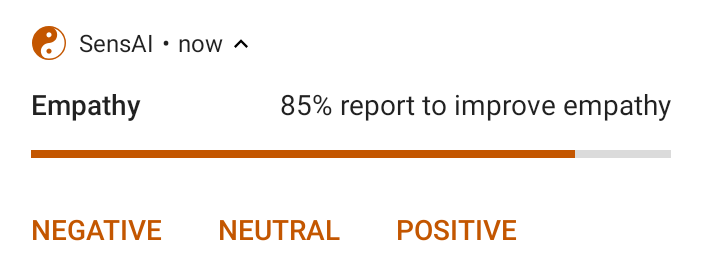}{Empathy notification including 
valence report buttons}

\texttt{Sentiment} analysis utility including integration with language
detection, translation and more is provided by specific libraries and services
included in \sensai. All contributing for the best effort to get the sentiment
value along with the language. A heuristic is in place to adapt the analysis to
human idiosyncratic aspects, such as mixed languages (English and Portuguese
supported) and emoticons amongst other abbreviations when writing short
messages. To deal with this rich and sometimes creative written content the best
effort approach is depicted in
Figure~\ref{fig:sensai---sentiment-heuristic-flow}.
\fig{scale=0.8}{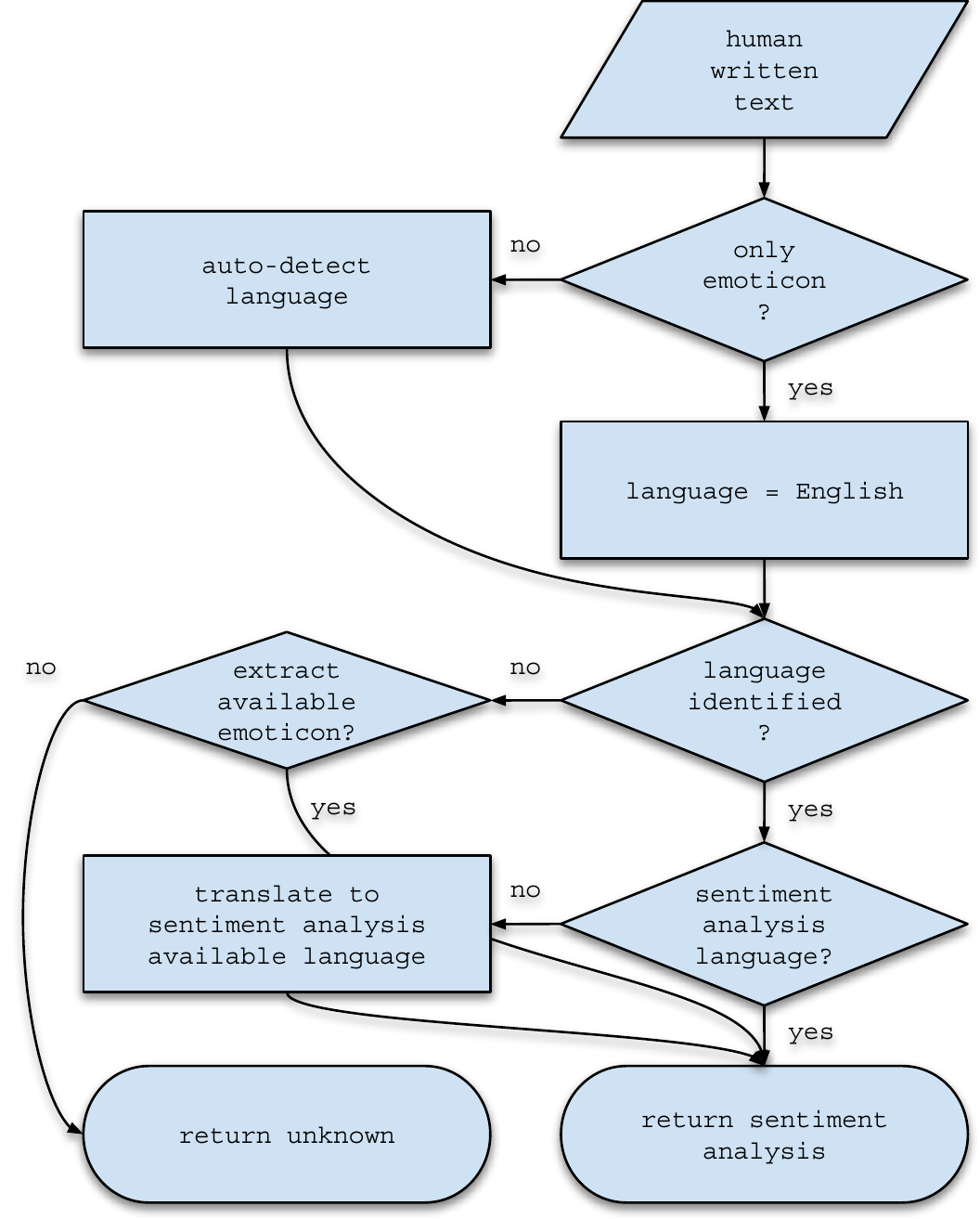}{Sentiment analysis heuristic}

\texttt{Expanse} is a periodic and scheduled service for data syncing with a
memory aggregator in the Cloud --- \sensaiexpanse. It is robust to failures
using a mechanism similar to a transaction, i.e., only successfully transferred
data is marked as such (able to be deleted after cache persistence time limit).
Moreover, on lack of a suitable data connection available it will adapt by
increasing verification frequency for later try to sync. This mechanism of
local cache and Cloud sync is paramount to restrict memory resources consumption
and guarantee proper data collection.

\subsection{\sensaiexpanse}

The agent Cloud-expanded resources --- \expanse\ --- are the augmentation spread
of the \sensai\ limited smartphone resources (e.g., data persistence, processing
and power). Stores data from all \sensai\ agents anonymously to guarantee that
the human's privacy is kept when the data flows to analysis. Includes repository
with historical data, processing algorithms, services of machine learning
towards prediction of emotional valence in context, i.e., \sensai\ augmented
memory and cognition. Moreover, processing all eligible data through available
algorithms towards \gls{automl} requires (a) adaptation to the diverse human
behaviours reflected in the data set; and (b) Bayesian efficient auto discovery
on parameters. Software architecture modules and services are depicted in
Figure~\ref{fig:expanse---software-architecture-modules}.

\fig{scale=0.54}{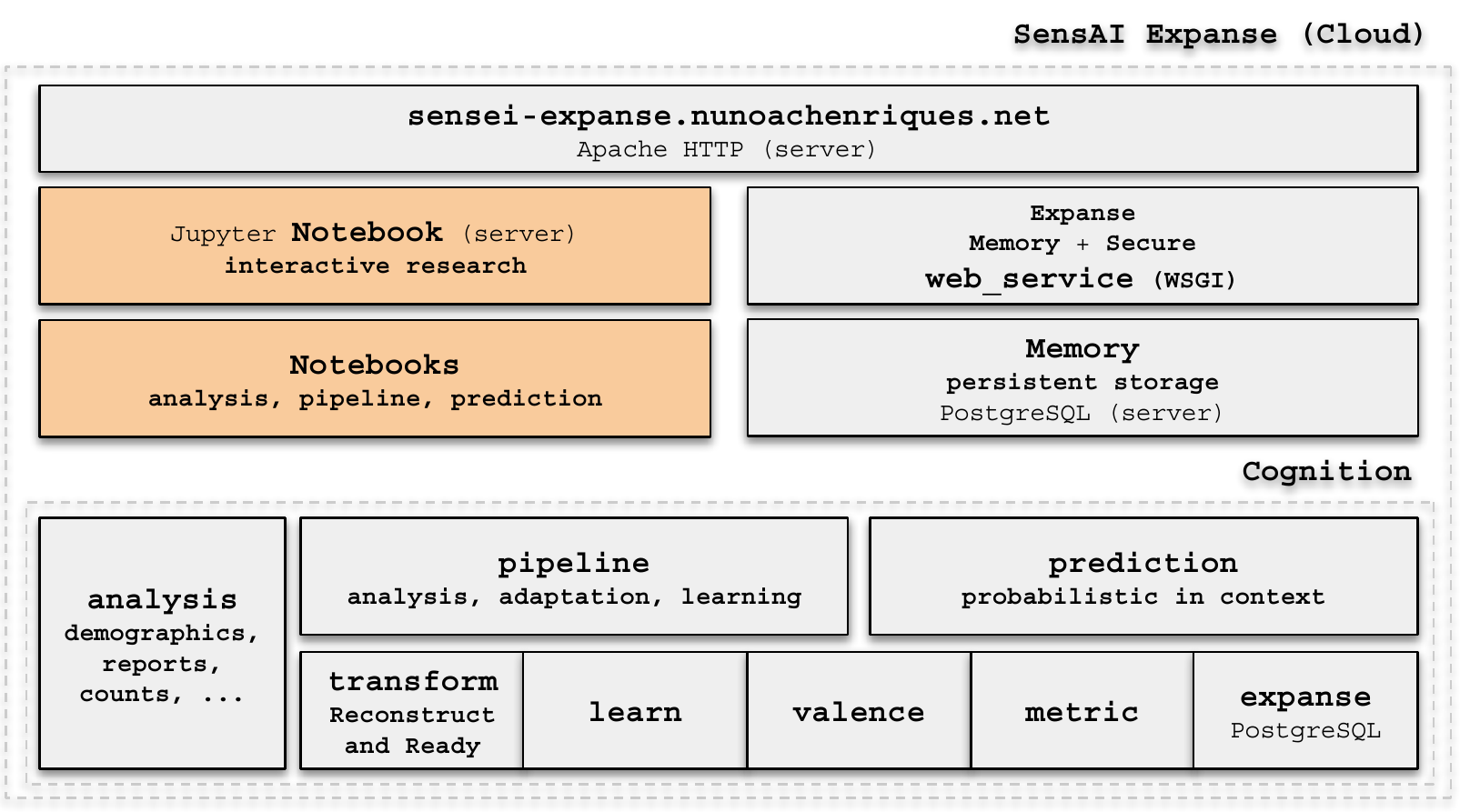}{\expanse\ software 
architecture modules and services}

Adaptive actions start, amongst other things, with \texttt{analysis} on
gathering data aggregations and filtering eligible human entities. This
eligibility selection has more steps through the \texttt{pipeline} process until
reaching the final data samples for machine learning. Before the final step,
\texttt{transform} acts on cleaning, reconstructing and fixing collected data
such as upsampling data within proper boundaries related to collecting
parameters previously used to save resources in \sensai. The \expanse\ developed
custom \texttt{pipeline} for \sensai\ learning uses a myriad of heuristics and
other algorithms. These include a data class (negative, neutral, positive)
imbalance (reports count) degree from \cite{Zhu2018}. Also, a custom valence
class count check and restrict in order to adapt the learning process in cases
such as emotional valence reported for only two (or even one) classes. The final
eligible entities are achieved after these valence class count and imbalance
degree processing.

The \texttt{learn} module integrates several state-of-the-art algorithms from
two main categories of (a) unsupervised ones, such as HDBSCAN for clustering
location coordinates and accurately drop outliers; and (b) supervised for
multi-class classification, such as \xgboostname\ by XGBoost and a custom
multilayer Perceptron using Keras in TensorFlow.

Additional steps are in place towards \gls{automl} by making use of (a) the
learn process calling a function for running HDBSCAN clustering algorithm on the
different \texttt{min\_samples} provided in order to find the best
\texttt{min\_cluster\_size} parameter; and before each call to one of the
classification algorithms learning process (b) an auto search is in place for
the best cross validation $N$ splits regarding the algorithm minimum number of
accepted classes. Next, a hyperparameter auto tuning with cross validation for
each specific model uses Bayesian optimisation. Finally, the model with
parameters fit for each human current data is achieved and performance metrics
such as F1 score are computed. The current knowledge from the learning process
is stored using the \texttt{expanse} module and the \texttt{Memory} component.
The \texttt{prediction} module is serving answers from Web service requests in
the Cloud. These responses are for contextualised (location and moment)
emotional valence prediction requests for the requiring human, as depicted in 
Figure~\ref{fig:sensai+expanse---secure-flow}.

\fig{scale=0.8}{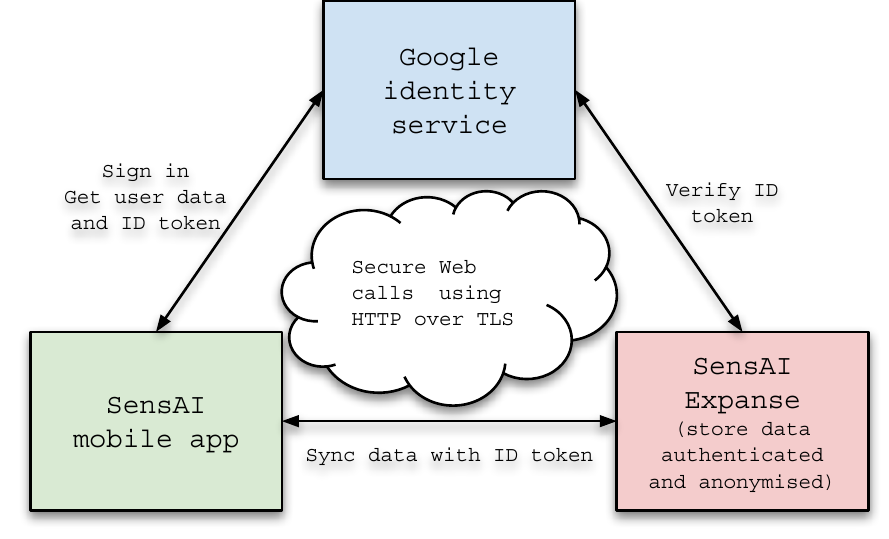}{End-to-end secure communication}

All communications are end-to-end secured and digitally signed, restricted to 
the owner available data, i.e., a human A cannot obtain the prediction for a 
human B.

\section{Study}\label{sec:study}

In this section, it will be described the applied method and the outcomes of a 
research study comprising a population of human individuals interacting with 
\sensai\ in the wild.

\subsection{Method}

The participants are gathered from all kinds and creeds, i.e., avoiding the
laboratory usual limitations, such as sampling only from \gls{weird} societies
known as a frequent bias \cite{Henrich2010}. This goal is accomplished by
choosing to collect data using smartphone sensing \cite{Cornet2018} by means of
an \android\ application. \sensai\ has already been installed by users from ten
different countries and four continents (Africa, America, Asia, Europe). A total
of 57 participants installed \sensai, eight were discarded for not sharing
demographic data thus 49 (18 female, 31 male) remained eligible. The pipeline
process further reduces the population to 31 eligible individuals after valence
class imbalance restrictions are applied. Moreover, demographic data comprises
birthdate and gender at this stage, extending the collection to other aspects
such as income and education level is foreseen as of interest for future
studies.

Regarding the system prediction performance, a comparison study assessing a few 
state-of-the-art algorithms and two metrics was in place. The set of estimators 
comprises linear (\lrname), non-linear (\xgboostname), and connectionist 
(\nnname) distinct approaches. Also, one more estimator is used as baseline 
(\dummyname) generating predictions by respecting the training set's class 
distribution (option \texttt{strategy=stratified}). The two metrics studied 
were F1 score (option \texttt{average=weighted}) and \gls{mcc} prepared for 
multi-class ($n=3$) case from \sklearnpkg\ package. Both metrics are very 
popular and with good software support for machine learning.

\subsection{Results}

The preliminary results were achieved by first using the \gls{mcc} metric on 
four estimators. The performance statistics depicted in 
Figure~\ref{fig:prediction---mcc-normal-model} revealed unexpected overly high 
values (e.g., baseline median near $0.8$) raising red flags about possible 
issues, such as estimators overfitting.

\fig{scale=0.4}{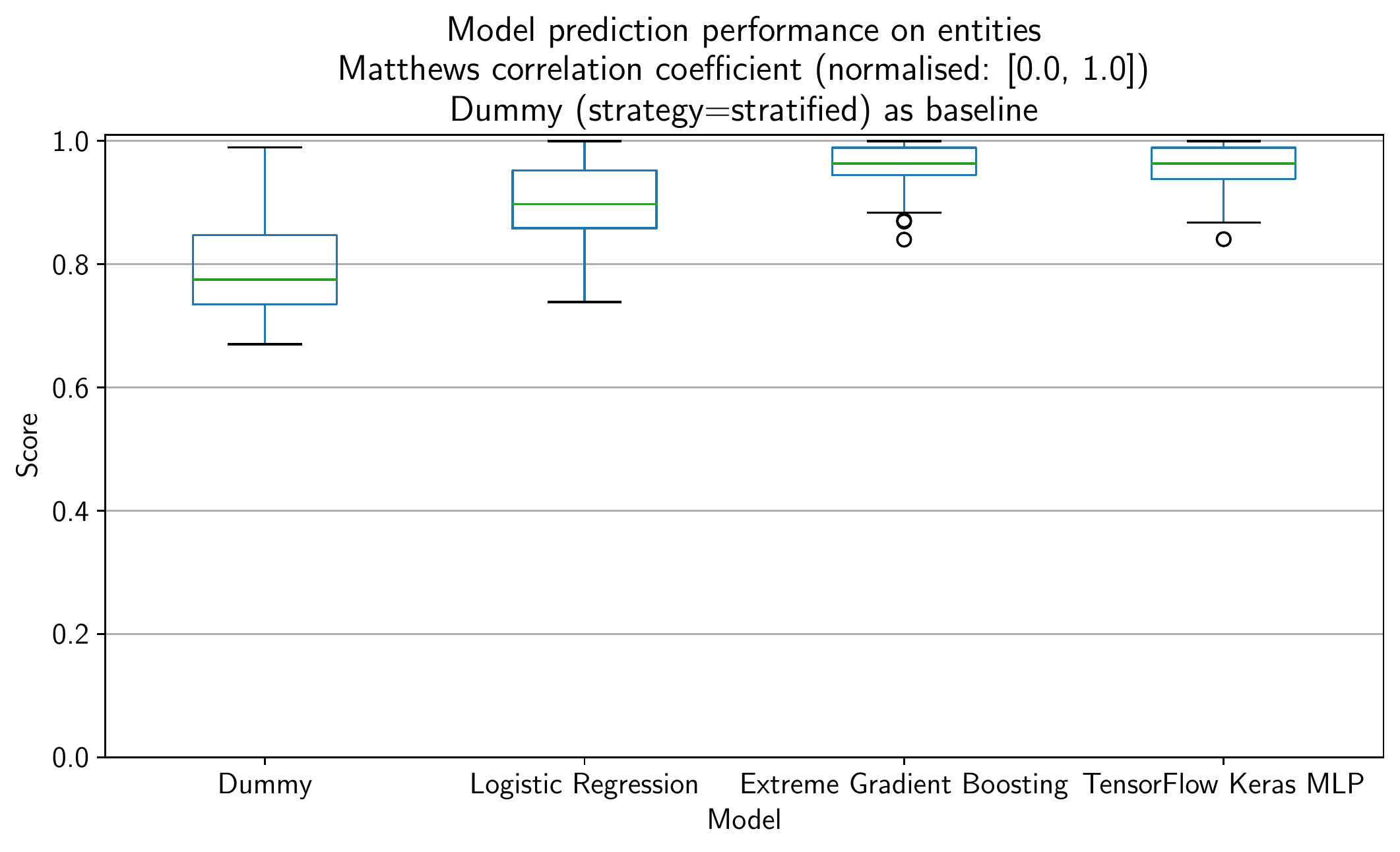}{Model prediction performance 
statistics using \gls{mcc}}

In order to verify the overfitting issue possibility on the first results (after
several runs to sustain the achieved values) a different score metric is used.
F1 score is selected for performance evaluation within the same experimental
conditions. The results are depicted in Figure~\ref{fig:prediction---f1-model}.

\fig{scale=0.4}{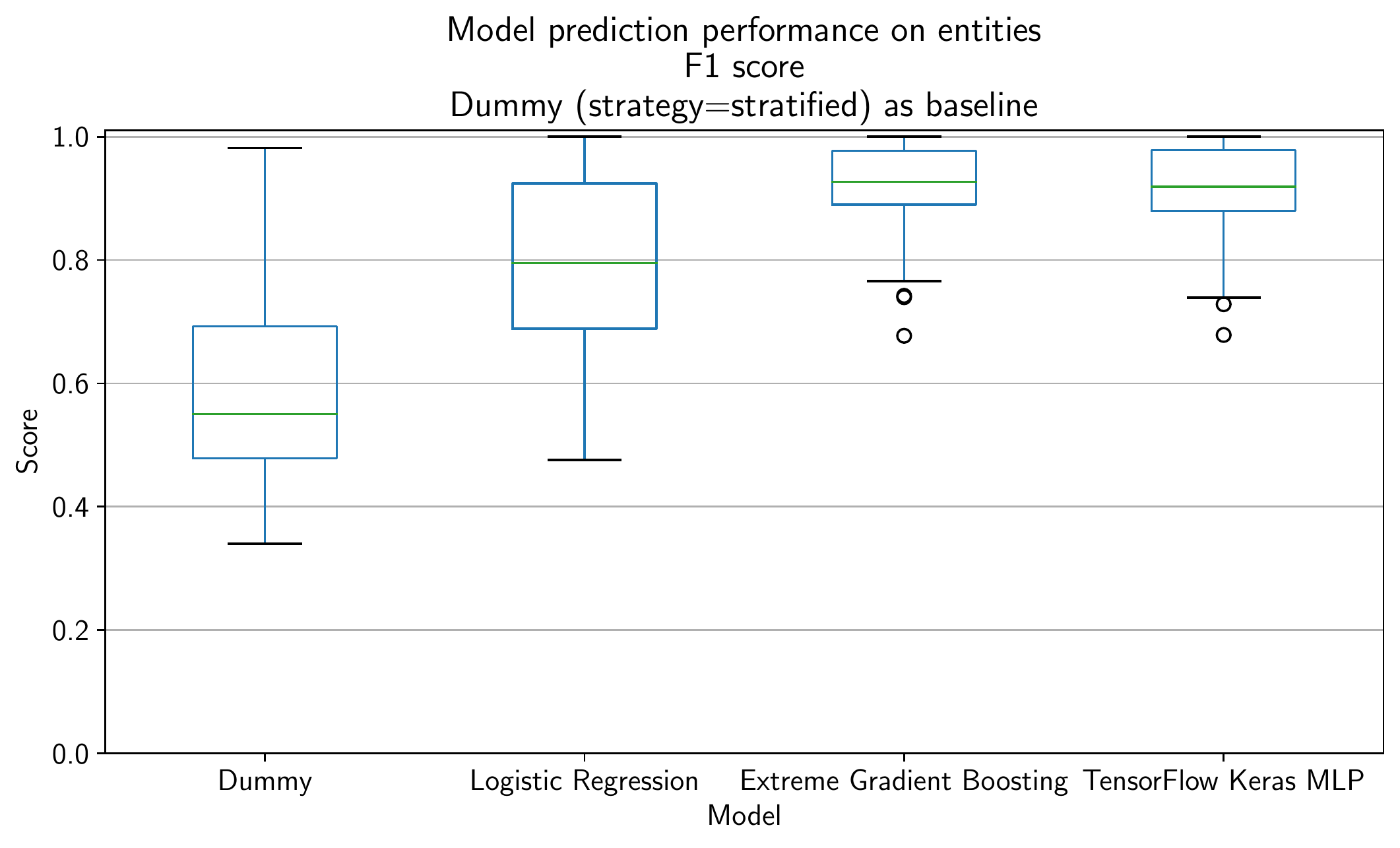}{Model prediction performance statistics 
using F1 score}

There is evidence of measurement discrepancies between F1 versus \gls{mcc} over
the population datasets, as depicted in
Figure~\ref{fig:prediction---f1-vs-mcc-dummy}. In order to assess the
significance of these differences, a Mann-Whitney U test is applied on F1 versus
\gls{mcc} and the results presented in Table~\ref{tab:utestf1vsmcc} show
that the null hypothesis ($H_0$: two sets of measurements are drawn from the
same distribution) can be rejected, i.e., evidence of significant differences on
F1 versus \gls{mcc} results.

\fig{scale=0.4}{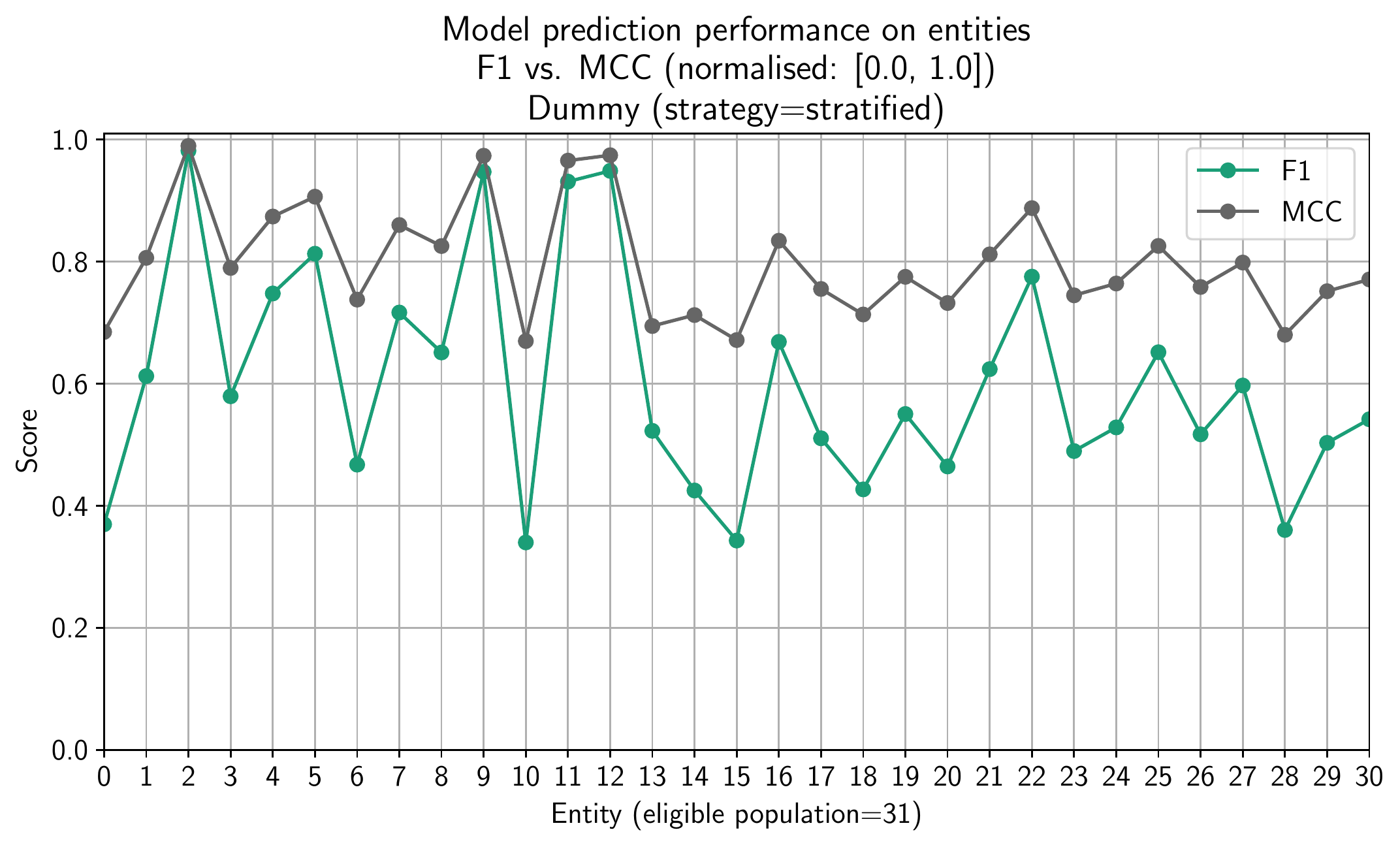}{Model prediction performance on 
entities: F1 vs. \gls{mcc}}

\captionsetup{font={footnotesize,sc},justification=centering,labelsep=period}%
\begin{table}[htbp]
\caption{F1 vs. \gls{mcc} (Figure~\ref{fig:prediction---f1-vs-mcc-dummy}): 
Mann-Whitney U test results}
\label{tab:utestf1vsmcc}
\centering%
\begin{tabular}{lS[table-format=1.3,table-figures-exponent=2,table-sign-exponent]c}
\toprule
Metrics & \pvalue & Meaning ($\alpha=0.05$)\\
\midrule
F1 vs. \gls{mcc} & 2.237e-02 & $H_0$ can be rejected ($p<\alpha$)\\
\bottomrule
\end{tabular}
\end{table}
\captionsetup{font={footnotesize,rm},justification=centering,labelsep=period}%

As a result, F1 measurements seem reasonably nearer to the expected baseline
statistics than \gls{mcc}. Further inspection using the confusion matrix for
some cases from the $26/31$ ($83\%$) with more than $10$ percentage points
difference uphold F1 score as more accurate than \gls{mcc}. Moreover, making use
of the classification report no pattern was identified correlating the inspected
cases, i.e., all distributions have distinct data shapes. Although the \gls{mcc}
is considered to have advantages over the F1 score, specifically in the binary
classification case, such claim \cite{Chicco2017} was not evidenced in this
research with the current eligible population and datasets. The results of this
study using F1 score for each model are summarised in
Table~\ref{tab:model-score-duration} and depicted in
Figure~\ref{fig:prediction---f1-model}.

\captionsetup{font={footnotesize,sc},justification=centering,labelsep=period}%
\begin{table}[htbp]
\caption{All four estimators average F1 score and total duration}
\label{tab:model-score-duration}
\robustify\bfseries%
\centering%
\begin{tabular}{lS[detect-weight]c}
\toprule
Model & {F1 average} & Duration\\
\midrule
\dummyname   &           0.600 &           00:00:46\\
\lrname      &           0.795 &           01:46:22\\
\xgboostname & \bfseries 0.910 & \bfseries 01:54:34\\
\nnname      &           0.907 &           24:35:11\\
\bottomrule
\end{tabular}
\end{table}
\captionsetup{font={footnotesize,rm},justification=centering,labelsep=period}%

The \xgboostname\ model is elected as best option amongst the other two
additional models plus baseline also available. The choice for this model is
further supported by an interest towards (a) \gls{xai} predictions; and (b)
efficient energy use besides overall score achievement. Regarding (a) the
\xgboostname\ includes feature importance scores for each entity model thus
proper explainable context (e.g., the specific location feature with the highest
score). As for (b) evidence already presented in
Table~\ref{tab:model-score-duration} show \xgboostname\ as the best performer
although marginally to the second yet for less than a tenth of the processing
duration. Therefore, \xgboostname\ is simultaneously on average the best model
for prediction with efficient energy use and also easy explainable by feature
importance inspection.

\section{Conclusion and Future Work}\label{sec:conclusion}

This paper has described the \sensaiplusexpanse\ smartphone sensing-based
system, distributed in the Cloud, robust to unexpected behaviours from humans
and the mobile demanding environment, able to predict emotional valence states
with very good performance. The \sensai\ agent adapts to the restricted
resources and volatile environment of a mobile device where an operating system
dictates behaviour rules. Any ill-behaved application is automatically stopped
by the system and may even be excluded from restarting. These cases where
processing takes too long usually result in the application being stopped and
declared \gls{anr}. There is no evidence of any \gls{anr} in the Google Play
Console used to monitor all events from \sensai. Conversely, there is evidence
of a few ``Illegal State'' crashes from which the agent recovered and continued
to interact after a maximum of fifteen-minute delay (interval for periodic
scheduled checks). Furthermore, battery consumption is kept at one digit
percentage (e.g., $1\%$) for a day-long use in several devices laboratory and
regular testing. The outcomes presented show evidence, restricted to population
and data samples in this research, of \sensaiplusexpanse\ ability to adapt and
learn to predict emotional valence states with a high score of $F1=0.910$ on
average (Table~\ref{tab:model-score-duration} and
Figure~\ref{fig:prediction---f1-model}). Therefore, \sensaiplusexpanse\
contributes as a novel platform for studies about human emotional valence
changes in context of location and moment. Moreover, it reinforces smartphone
sensing contribution as a tool for continuous, passive, and personalised health
check, such as emotional disturbances, in spatial and temporal context.
Furthermore, all the source code is published as free software under the Apache
License 2.0. Future work should investigate emotional valence report
discrepancies amongst population demographics, such as age and gender.
Furthermore, an assessment over the agent robustness to those differences would
be of interest. Thus, studies about the agent's neutrality to distinct age
ranges and gender combinations should be in place by means of the elected
\xgboostname\ model with F1 score.
 
\section*{Acknowledgement}

N.A.C.H. thanks Jorge M. C. Gomes for the precious contributions. This work is
partially supported by \textit{Universidade de Lisboa} [PhD support grant
May 2016--April 2019]. Partially supported by
\textit{Funda\c{c}\~{a}o para a Ci\^{e}ncia e Tecnologia} [UID/MULTI/04046/2019
Research Unit grant from FCT, Portugal (to BioISI)]. This work used the European
Grid Infrastructure (EGI) with the support of NCG-INGRID-PT.

\bibliographystyle{IEEEtranNACH}\bibliography{sensai-adaptive}
\end{document}